\documentclass[10pt,twocolumn,letterpaper]{article}

\usepackage[pagenumbers]{wacv} 

\usepackage{graphicx}
\usepackage{bm}
\usepackage{algorithm}
\usepackage{algpseudocode}
\usepackage{mathtools}
\usepackage{multirow}
\usepackage{tabularx}
\usepackage{pifont}
\newcommand{\fst}{\textbf}
\newcommand{\scd}{\textun}

\newcommand{\textun}[1]{\underline{#1}}
\newcommand{\cmark}{\ding{51}}
\newcommand{\xmark}{\ding{55}}

\def\eqref#1{equation~\ref{#1}}

\def\1{\bm{1}}

\newcommand{\too}{\rightarrow}

\def\vzero{{\bm{0}}}

\def\vtheta{{\bm{\theta}}}

\def\vr{{\bm{r}}}

\def\vx{{\bm{x}}}

\def\vz{{\bm{z}}}

\def\mI{{\bm{I}}}

\DeclareMathAlphabet{\mathsfit}{\encodingdefault}{\sfdefault}{m}{sl}
\SetMathAlphabet{\mathsfit}{bold}{\encodingdefault}{\sfdefault}{bx}{n}

\def\gN{{\mathcal{N}}}

\def\gU{{\mathcal{U}}}

\def\sR{{\mathbb{R}}}

\newcommand{\E}{\mathbb{E}}

\DeclareMathOperator*{\argmin}{arg\,min}

\definecolor{wacvblue}{rgb}{0.21,0.49,0.74}
\usepackage[pagebackref,breaklinks,colorlinks,allcolors=wacvblue]{hyperref}

\def\wacvPaperID{2084} 
\def\confName{WACV}
\def\confYear{2026}

\title{ODE$_t$(ODE$_l$): Shortcutting the Time and the Length in \\ Diffusion and Flow Models for Faster Sampling}

\author{Denis~Gudovskiy\textsuperscript{\rm 1}\thanks{Corresponding author: \texttt{denis.gudovskiy@us.panasonic.com}}
	\quad~Wenzhao~Zheng\textsuperscript{\rm 2}
	\quad~Tomoyuki~Okuno\textsuperscript{\rm 3}
    \quad~Yohei~Nakata\textsuperscript{\rm 3}
    \quad~Kurt~Keutzer\textsuperscript{\rm 2}\\
	{\textsuperscript{\rm 1}Panasonic AI Lab} ~~
    {\textsuperscript{\rm 2}UC Berkeley} ~~
    {\textsuperscript{\rm 3}Panasonic DX-CPS}
}

\begin{document}

\maketitle

\begin{abstract}
Continuous normalizing flows (CNFs) and diffusion models (DMs) generate high-quality data from a noise distribution. However, their sampling process demands multiple iterations to solve an ordinary differential equation (ODE) with high computational complexity.
State-of-the-art methods focus on reducing the number of discrete time steps during sampling to improve efficiency.
In this work, we explore a complementary direction in which the quality-complexity tradeoff can also be controlled in terms of the neural network length.
We achieve this by rewiring the blocks in the transformer-based architecture to solve an inner discretized ODE \wrt its depth.
Then, we apply a length consistency term during flow matching training, and as a result, the sampling can be performed with an arbitrary number of time steps and transformer blocks.
Unlike others, our ODE$_t$(ODE$_l$) approach is solver-agnostic in time dimension and reduces both latency and, importantly, memory usage. 
CelebA-HQ and ImageNet generation experiments show a latency reduction of up to $2\times$ in the most efficient sampling mode, and FID improvement of up to $2.8$ points for high-quality sampling when applied to prior methods. We open-source our code and checkpoints at \href{https://github.com/gudovskiy/odelt}{github.com/gudovskiy/odelt}.
\end{abstract}

\section{Introduction}\label{sec:intro}
Recent flow matching (FM) for continuous normalizing flows (CNFs) \citep{lipman2023flow, albergo2023building, liu2023flow} and previously explored diffusion models (DMs) \citep{sohl2015diffusion, song_generative_2019, ho_denoising_2020} have been successful in modeling various data domains, including images \citep{esser2024scaling}, videos \citep{bar2024lumiere}, language \citep{sahoo2024simple}, multimodal data \citep{omniflow}, proteins \citep{geffner2025proteina} \etc

\begin{figure}
\includegraphics[width=0.65\columnwidth]{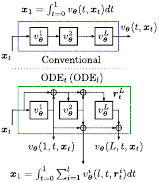}
\centering
\caption{The conventional approach (top) models the interpolated vector field $u_t (\vx | \vz)$ by an expressive but monolithic neural network $v_\vtheta (t, \vx_t)$. This limits practitioners to adjust the quality-complexity tradeoff only in the integral's time dimension. Our ODE$_t$(ODE$_l$) approach (bottom) considers the neural network as the inner $\textrm{ODE}_l$ and allows to select the number of active blocks and, hence, reduce latency and memory usage during sampling.}
\label{fig:problem}
\end{figure}

\citet{tong2024improving} have generalized the training phase in such models with the conditional FM (CFM) framework using the simulation-free regression objective. In addition, \citet{song2021scorebased} showed that the stochastic differential equation (SDE) to sample from the DM can be converted to a probability flow ordinary differential equation (ODE) that further unifies two types of generative models. Despite simplified training, CNFs and DMs have the major bottleneck in the sampling process where the ODE needs to be solved with a large number of function evaluations (NFE).

To improve efficiency, most existing methods focus on the minimization of NFE to decrease the number of ODE integration steps when sampling data from the noise distribution \citep{shen2025efficient}. \citet{dieleman2024distillation} argues that, although practitioners could benefit from NFE reduction methods, \textit{there is no free lunch and these methods are inherently limited by the quality-complexity tradeoff}. At the same time, a recent class of such methods \eg shortcut \citep{shortcut}, inductive moment matching \citep{imm} and mean flow models \citep{mf} provide the important property to select the NFE budget while significantly minimizing the drop in sampling quality.

In this paper, we propose a complementary approach to further reduce test-time computations in CNFs and DMs. Unlike recent methods \citep{shortcut,imm,mf} that only minimize the NFE, we are motivated to modify the time-dependent neural network that approximates the target vector field. In particular, we consider an architecture that consists of a sequence of blocks \eg the transformer-based DiT \citep{dit} or SiT \citep{sit}, where the number of active blocks can be dynamically changed. Then, the quality-complexity tradeoff also depends on the selected neural network depth and scales better than the NFE-only baseline in Figures~\ref{fig:qual}-\ref{fig:scaling}. We show that our objective can be achieved with minor architectural modifications and negligible computational overhead.

Our \textit{ODE within ODE approach}, dubbed ODE$_t$(ODE$_l$), is sketched in Fig. \ref{fig:problem}.  
Unlike conventional architectures that treat the neural network $v_\vtheta (t, \vx_t)$ as a monolithic function, we consider it as a compositional function. When implemented as a sequence of blocks, it can solve an \textit{inner discrete ODE} \wrt the sequence length. In particular, we condition the modeled vector field on the length $l$ that defines the number of active blocks to produce a length-dependent inner ODE solution $v_\vtheta (l, t, \vx_t)$. Then, it is used in the \textit{outer ODE} integration \wrt the time variable. For example, our neural network has the original full length when $l=L$. On the other hand, our method reduces the computational complexity with $l<L$ at the expense of sampling quality. Our main contributions are summarized as follows:
\begin{itemize}
\item Our ODE$_t$(ODE$_l$) extends NFE minimization methods to the neural network length dimension for CNFs and DMs.
\item It is uniquely solver-agnostic in the outer ODE$_t$, reduces latency and mitigates the memory bottleneck in the inner $\textrm{ODE}_l$ with a negligible architectural overhead.
\item Experiments show further improvement in the quality-complexity tradeoff scaling when the depth of the neural network is modulated using the proposed ODE$_t$(ODE$_l$).
\end{itemize}

\section{Related work}\label{sec:related}
We overview key concepts that have been used to improve efficiency in a class of generative models~\ie~the CNFs and DMs. Then, we introduce up-to-date NFE minimization related work and present a broader context with developments in architecture- and solver-level optimizations.

\textbf{Knowledge distillation}. Knowledge distillation for discriminative models has been pioneered by \citet{hinton2015}, who showed that a larger "teacher" network knowledge can be transferred to a smaller "student" one by mimicking the logit outputs without a substantial drop in speech and image recognition accuracy. An extension of this approach for generative CNFs and DMs has been realized by distilling these models \wrt the discretized time. With this approach, only a single time step \citep{luhman2021knowledge} or a few steps \citep{salimans2022, Meng_2023_CVPR} of the student can be enough to mimic the teacher's ODE solution with reasonable precision. We also highlight SlimFlow \citep{zhu2025slimflow} with an explicit objective to distill to a smaller student network. Unlike SlimFlow, our ODE$_t$(ODE$_l$) architecture modifications lead to a number of sub-networks after the end-to-end training, where these sub-networks can also be considered as a special type of implicit student networks.

\textbf{Rectified flows and consistency modeling}. \citet{cm} propose a novel type of distillation without an explicit student network, and a one-step model can be trained end-to-end from scratch. During consistency training, the neural network is constrained to be \textit{self-consistent}~\ie~points on the same ODE sampling trajectory map to the origin. Later, consistency models have been improved \citep{song2024improved}, simplified \citep{geng2025consistency} and extended to the latent space \citep{luo2023latent}. 

Consistency is also related to the flow matching framework \citep{lipman2023flow, albergo2023building} with straight paths \eg rectified flows \citep{liu_rectified_2022}. Compared to diffusion models, rectified flows can achieve higher sampling quality with fewer integration steps. This has been explored in Reflow \citep{liu2023flow} with the NFE minimization procedure. \citet{shortcut} propose to employ both self-consistency and straight trajectories, where the shortcuts in the rectified flow provide dynamic selection of $\log_2$ discrete sampling steps with lower quality overhead. Recent inductive moment matching \citep{imm} and mean flow methods \citep{mf} introduce novel neural network parameterizations and training objectives that further improve one- and few-step generation quality. Our method is complementary to the above NFE optimization methods and, when combined, results in better quality-complexity scaling.
  
\textbf{Efficient architectures}. Architecture-level optimizations are typically explored separately from ODE acceleration. At the same time, minimization of the NFE is directly related to the complexity of the underlying neural network architectures. Earlier research has been proposed for discriminative models, including the concept of conditional computations \citep{bengio2016} and its realizations in architectures that can dynamically drop \citep{Wu_2018_CVPR} or skip \citep{Wang_2018_ECCV} blocks in neural networks. Similarly, early exit strategies have been proposed for transformers \citep{schuster2021} and more recently for DMs \citep{MoonCYYLC024}. Moreover, \citet{zhao2025dynamic} propose dynamic diffusion transformers for the DiT architecture \citep{dit} with dynamic tokens in time and spatial dimensions during generation. Similarly, \citet{you2025layer} introduce the dynamic architecture for DMs with efficient token-level routing mechanism.

\textbf{Alternative architectures and solvers}. Our method that modifies a vanilla transformer architecture is also analytically related to alternative architecture types. For example, recent fixed-point diffusion models \citep{Bai_2024_CVPR} are derived from deep equilibrium models (DEQs) \citep{bai2019deep} where each layer has infinite depth. In contrast to DEQ models, we consider each DiT or SiT block as an ODE step for the fixed-step Euler solver rather than the implicit DEQ layers with non-ODE solvers \citep{torchdeq}. Therefore, we rely on an ODE-centric architecture that has also been explored in multistep ResNet-type architectures for classifiers \citep{beyond, anode}, DMs \citep{Zhang_2024_CVPR}, neural flows \citep{bilos}, neural residual DMs \citep{ma2024neural} and PDE-Nets \citep{pdenet}. Lastly, there is a line of research \citep{zheng2023dpmsolverv, Zhou_2024_CVPR, chen2024on, frankel2025ss, wang2025differentiable} that introduces advanced ODE$_t$ solvers that select or learn more efficient sampling trajectories in the time dimension. In contrast, ODE$_t$(ODE$_l$) advances sampling efficiency by focusing on the inner ODE$_l$, which effectively represents active neural network layers when sampling from CNFs and DMs.

\section{Preliminaries}\label{sec:theory}
\textbf{Continuous normalizing flows.} Our notation follows \citep{lipman2023flow, tong2024improving}. There are a pair of data distributions $q(\vx_0)$ and $q(\vx_1)$ over $\sR^D$ with densities $p(\vx_0)$ and $p(\vx_1)$, respectively. Typically, $p_0=p(\vx_0)$ represents a known prior distribution. The task is to sample $\vx_1 \sim q(\vx_1)$ with unknown data density $p_1=p(\vx_1)$ and only access to an empirical distribution $\hat{q}(\vx_1)$ \ie the training set.

Formally, there are a \emph{probability density path} $p:[0,1]_t \times \sR^D \too \sR_{\ge 0}$, which is a time-dependent probability density function $p_t(\vx)$ with $t\in [0,1]$ such that $\int p_t(\vx)dx = 1$, and a time-dependent Lipschitz-smooth \emph{vector field} $u : [0,1]_t \times \sR^D \too \sR^D$. The vector field $u_t$ is used to construct a time-dependent diffeomorphism \ie, the CNF $\phi:[0,1]_t \times \sR^D \too \sR^D$ that is defined via the ODE as
\begin{equation}\label{eq:cnf}
d\phi_t(\vx)/dt = u_t(\phi_t(\vx))~\textrm{and}~ \phi_0(\vx) = \vx_0,
\end{equation}
where $\phi_t(\vx)$ is the ODE solution with $\phi_0(\vx)$ initial condition that transports $\vx$ from time $0$ to time $t$.

The vector field $u_t(\phi_t(\vx))$ is often modeled without the diffeomorphism's invertability property by an arbitrary neural network $v_\vtheta (t, \vx_t)$ with learnable weights $\vtheta$. 

\textbf{Flow matching training.} CNF training using the maximum likelihood objective with integration is computationally expensive \citep{ffjord, aca}. The FM framework \citep{lipman2023flow} proposes an alternative objective that regresses $v_\vtheta (t, \vx_t)$ to $u_t$ by conditioning the latter on a vector $\vz=\vx_1$. This has been extended by the conditional FM (CFM) framework \citep{tong2024improving} where $u_t(\vx | \vz)$ and $p_t(\vx | \vz)$ are conditioned on a more general $\vz \sim q(\vz)$. Then, the Gaussian conditional probability path and a unique target conditional vector field are written by
\begin{equation}\label{eq:cpath_cvec}
\begin{split}
p_t(\vx \vert \vz) & = \gN(\vx \, \vert\, \mu_t(\vz), \sigma_t(\vz)^2 \mI ), \\ u_t(\vx \vert \vz) & = \left( \vx - \mu_t(\vz) \right) \sigma'_t(\vz) / \sigma_t(\vz) + \mu'_t(\vz),
\end{split}
\end{equation}
where $\mu_t(\vz)$ and $\sigma_t(\vz)$ are the mean and standard deviation of the Gaussian distribution, respectively.

As a result, the CFM regression objective for simulation-free CNF training using (\ref{eq:cpath_cvec}) can be expressed as
\begin{equation}\label{eq:cfm}
\argmin\nolimits_{\vtheta} \mathcal{L}_\textrm{CFM} := \E_{t, \vx_t, \vx, \vz} \| v_\vtheta(t, \vx_t) - u_t(\vx|\vz) \|^2,
\end{equation}
where the time $t \sim p(t)$ is sampled from the uniform or $\log$-normal distribution \citep{esser2024scaling}, and the trajectory $\vx_t = \mu_t(\vz) + \sigma^2_t(\vz) \vx$ is defined by the selected conditional distribution $q(\vz)$ and the standard Gaussian $\vx \sim \gN(\vzero,\mI)$.

Widely-used rectified flow matching \citep{liu_rectified_2022} with the linear deterministic interpolant sets $q(\vz)=q(\vx_0) q(\vx_1)$ with $\mu_t(\vz)=t \vx_1 + (1-t) \vx_0$ and $\sigma^2_t(\vz)=0$. The CFM framework in (\ref{eq:cfm}) can be extended to diffusion models by choosing an appropriate stochastic interpolant. For example, the variance preserving DM \citep{impr_ddpm} employs $q(\vz)=q(\vx_1)$, $\mu_t(\vz)=\alpha_{1-t} \vx_1$ and $\sigma^2_t(\vz)=1- \alpha_{1-t}^2$. The ODE perspective (\ref{eq:cnf}) for DMs \citep{song2021scorebased} is particularly useful for the faster deterministic sampling that has been discussed in \citep{zhang2023fast, lipman2023flow}.

\section{Proposed method}\label{sec:prop}
\textbf{Problem statement for data sampling.} Although (\ref{eq:cfm}) avoids integration during the training phase, the sampling process still requires to solve the ODE in (\ref{eq:cnf}). In general, the test-time sampling process can be written as
\begin{equation}\label{eq:sample}
\frac{d\phi_t(\vx)}{dt} = u_t(\phi_t(\vx))~\implies~\vx_1 = \int_{0}^{1} v_\vtheta (t, \vx_t) dt,
\end{equation}
where the initial $\vx_0$ is sampled from the known prior $q(\vx_0)$.

\textbf{Compositionality as the inner ODE$_l$.} Conventional architectures that implement the neural network $v_\vtheta (t, \vx_t)$ in (\ref{eq:cfm}) consider the diffeomorphism $\phi_t(\vx)$ as a single and expressive time-dependent function. However, widely-used neural networks have layered architectures, and, moreover, they are typically implemented as a sequence of blocks. With this observation, we propose to define (\ref{eq:sample}) as the \textit{outer} $\textrm{ODE}_t$ \wrt time $t$, whereas $v_\vtheta (t, \vx_t)$ itself can represent the solution to an \textit{inner} $\textrm{ODE}_l$ \wrt its depth or length $l$.

We convert conventional $\textrm{ODE}_t$ approach to the proposed ODE$_t$(ODE$_l$) in several steps.
First, we redefine $\phi_t(\vx)$ as a compositional function $\phi^L_t \circ \cdots \phi^l_t \cdots \circ \phi^1_t(\vx) \rightarrow \phi_t(\vx)$ with up to $L$ layers. 
Then, we introduce the length-dependent variable $l$ that activates only the first $l$ layers in the compositional function $\phi_t(\vx)$. In other words, the extended diffeomorphism is defined as $\phi:\{1,\ldots, L\}_l \times [0,1]_t \times  \sR^D \too \sR^D$. This assumption is valid for recent transformer-based architectures \citep{dit,sit}. Then, we rewrite the sampling (\ref{eq:sample}) as
\begin{equation}\label{eq:samplelt}
\frac{d}{dt} \left( \frac{\phi^l_t(\vx)}{dl} \right) = u_t \left(\phi^l_t(\vx) \right)~\Rightarrow~\vx_1 = \int_t \int_l v^l_\vtheta (t, \vx_t) \,dl\,dt.
\end{equation}

Although there are more efficient advanced solvers \citep{zheng2023dpmsolverv, frankel2025ss}, recent NFE minimization methods \citep{shortcut, imm, mf} rely on a low-complexity Euler solver to replace the outer ODE$_t$ integration in (\ref{eq:samplelt}) by a recurrent summation
\begin{equation}\label{eq:sumt}
\vx_{t+\Delta_t} = \vx_{t} + \Delta_t v_\vtheta (l, t, \vx_t)~\textrm{and}~\Delta_t = 1 / T,
\end{equation}
where $T$ is the number of discrete time steps.

\textbf{Architectural modifications.} Unlike the continuous time variable, the neural network's inner ODE$_l$ always has a discrete length. Hence, we apply the same first-order Euler solver from (\ref{eq:sumt}) to the neural network length variable in (\ref{eq:samplelt}). This step induces the proposed architectural modification with a network rewiring. To output the $\textrm{ODE}_l$ solution, we introduce residual connections $\vr^i_t$ that propagate latent vectors from layer to layer, and time- and length-dependent outputs of $v_\vtheta(l, t, \vx_t)$ as shown in Fig. \ref{fig:problem}. Then, our $v_\vtheta(l, t, \vx_t)$ represents a solution to the inner ODE$_l$ for the selected length $l$ from the $\{1,\ldots, L\}$ range. Overall, the inner ODE$_l$ processing at a time $t$ can be expressed as
\begin{equation}\label{eq:dode}
\begin{split}
v_\vtheta (l, t, \vx_t) := \sum\nolimits^{l}_{i=1} v^i_\vtheta (l, t, \vr^i_t), ~\vr^0_t = \vx_t,\\
\vr^{i}_t = \vr^{i-1}_t + \Delta_l v^i_\vtheta (l, t, \vr^{i-1}_t)~\textrm{and}~\Delta_l = 1 / l,
\end{split}
\end{equation}
where $\vr^{i}_t$ is the $i$-th layer residual and $\Delta_l$ is the step size.

\begin{figure*}[t]
\centering
\includegraphics[width=0.54\textwidth]{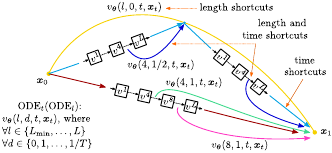}
\caption{\textbf{Visualization of ODE$_t$(ODE$_l$) when it is applied to the \citep{shortcut} time shortcuts.} Our approach models the target vector field using the configurable $v_\vtheta(l, d, t, \vx_t)$ neural network. The length hyperparameter $l$ defines the number of active blocks within the architecture \ie~length shortcuts. The time shortcuts are adjusted by the hyperparameter $d$ as proposed by \citet{shortcut}. Then, several cases can be highlighted: a) $v_\vtheta(L, 0, t, \vx_t)$ is equivalent to the conventional CFM/DM processing, b) $v_\vtheta(L, d, t, \vx_t)$ is identical to the time-only shortcuts (\eg, $d=1$ for single-step sampling), c) $v_\vtheta(l, 0, t, \vx_t)$ with only length shortcuts supports any ODE solver including more advanced ones with adaptive steps \citep{dormand1980family, zheng2023dpmsolverv, frankel2025ss}, and d) $v_\vtheta(l, d, t, \vx_t)$ is the general setup with length and time shortcuts.}
\label{fig:arch}
\end{figure*}

Our architectural extension (\ref{eq:dode}) with length shortcuts provides an additional knob to control the complexity-quality tradeoff of sampled data as shown in Fig.~\ref{fig:arch}. For example, our method can combine both time and length shortcuts or any combination of them. Also, this provides additional unique properties: $\textrm{ODE}_t$ is solver-agnostic and $\textrm{ODE}_{l<L}$ decreases both the latency and memory usage. Although we rely on the first-order Euler method in (\ref{eq:dode}), other solvers can also be explored for ODE$_l$.

\textbf{Training with length shortcuts.} In a third step, we describe the training procedure with length shortcuts in (\ref{eq:dode}) and impose a consistency objective on the outputs of the length-dependent network. Note that the time-dependent probability path and the vector field $u_t(\vx \vert \vz)$ from (\ref{eq:cpath_cvec}) have not been influenced by the change in the neural network $v_\vtheta(l, t, \vx_t)$ and have the same trajectories in the time dimension. Then, we follow the consistency training approach \citep{song2024improved} and introduce an additional length consistency term $(\mathcal{L}_\textrm{LC})$ to the optimization objective. This term depends on the full-length \textit{teacher} and the implicit \textit{students} with uniformly sampled lengths $l$ from the range $\{L_\textrm{min}, \ldots,  L\}$, where $L_\textrm{min}$ is the minimal network length. Formally, our training objective $\mathcal{L} = \mathcal{L}_\textrm{CFM} + \mathcal{L}_\textrm{LC}$ is defined as
\begin{equation}\label{eq:ours}
\begin{split}
\argmin\nolimits_{\vtheta} \mathcal{L} & := \E_{l \sim \gU \{ L_\textrm{min}, L \}} \|  v_\vtheta(l, t, \vx_t) - u_t(\vx|\vz) \|^2 \\
& +  \| v_\vtheta(l, t, \vx_t) - \operatorname{sg}(v_\vtheta(L, t, \vx_t)) \|^2,
\end{split}
\end{equation}
where $\operatorname{sg}$ is the stop-gradient operator, $t \sim p (t)$ and $\vx_t\sim p_t(\vx|\vz)$ are from the Section~\ref{sec:theory} framework.

Enforcing the self-consistency by bootstrapping the neural network targets in the (\ref{eq:ours}) second term has shown its efficiency in \citep{gu2023boot, shortcut}. This reduces the variance in the first empirical term and, instead, replaces it with deterministic neural network targets in the second term. To avoid a major increase in training time when calculating these targets, the mini-batch with the (\ref{eq:ours}) loss is constructed to contain only a small portion (\eg, $K=1/8$) of the deterministic consistency terms, while the rest of it contains empirical CFM terms. In general, our ODE$_l$ can be applied to any NFE optimization method \citep{shortcut, mf} that is shown in Algs.~\ref{alg:1}-\ref{alg:2}.

\begin{algorithm}[t]
\caption{ODE$_t$(ODE$_l$) training with NFE minimization baselines \citep{shortcut, mf} and our length shortcuts that are facilitated by the mini-batched length consistency (LC) term.}
\label{alg:1}
\begin{algorithmic}
\State \textbf{Input:} $q(\vz), p_t(\vx \vert \vz), u_t (\vx | \vz)$ for CFM/DM, $\vtheta \gets \vtheta_\mathrm{0}$
\While{training}
\State $\vx_0 \sim \mathcal{N}(0, \mI)$, $\vx_1 \sim \hat{q}(\vx_1)$, $\vz \sim q(\vz)$
\State $t \sim p(t)$, $\vz \sim q(\vz)$, $\vx_t \sim p_t(\vx \vert \vz)$~\Comment{Sec.~\ref{sec:theory}}
\State $l \sim \gU \{L_{\textrm{min}},L\}$~\Comment{length $l$ shortcuts}
\State $d \sim p(d) $~\Comment{NFE parameterization \citep{shortcut,mf}}
\State $v_\textrm{LC} \gets v_\vtheta(l, d, t, \vx_t)$~\Comment{length consistency students}
\State $u_\textrm{LC} \gets \operatorname{sg}(v_\vtheta(L, d, t, \vx_t))$~\Comment{stop gradient for teacher}
\State $v_\textrm{CFM+LC} \gets [v_\vtheta(l, d, t, \vx_t), v_\textrm{LC}]$~\Comment{$\operatorname{concat}$ predictions}
\State $u_\textrm{CFM+LC} \gets [u_t (\vx | \vz), u_\textrm{LC}]$~\Comment{$\operatorname{concat}$ targets}
\State $\vtheta \gets \operatorname{Update} \left(\vtheta, \nabla_\vtheta \| v_\textrm{CFM+LC} - u_\textrm{CFM+LC} \|^2 \right)$
\EndWhile
\end{algorithmic}
\end{algorithm}

\begin{algorithm}[t]
\caption{Efficient sampling with NFE minimization baselines \citep{shortcut, mf} in the outer $\textrm{ODE}_t$ using $d$ parameterization and our neural network shortcuts $l$ in the inner $\textrm{ODE}_l$.} 
\label{alg:2}
\begin{algorithmic}
\State \textbf{Input:} $q(\vx_0)$, $\vtheta$,
\State ~~~~~~~~~~~~$l \gets \{L_\textrm{min}, \ldots,  L\}$,
\State ~~~~~~~~~~~~$d \gets$ time shortcut \citep{shortcut} or integration interval \citep{mf}
\Function{$\operatorname{ODEsolve}_l$}{$t, \vx_t$}
\State $\vr^0_t = \vx_t$
\ForAll {$i \in \{1, \ldots, l\}$}
\State $\vr^{i}_t \gets \vr^{i-1}_t + \Delta_l v^i_\vtheta(l, d, t, \vr^{i-1}_t)$
\EndFor
\EndFunction\Comment{return $v_\vtheta (l, d, t, \vx_t)$}
\State $t \gets 0$, $\vx_t \sim q(\vx_0)$
\State $\vx_1 \gets \mathrm{ODEsolve}_t \mathrm{ODEsolve}_l(t, \vx_t)$\\
\Return{$\vx_1$}
\end{algorithmic}
\end{algorithm}

\textbf{Computational overhead.} We add a single learnable embedding for the proposed length $l$ shortcuts. The output of this embedding is added to the conventional positional, classifier-free label embeddings, and, if used, $d$ embeddings from \citep{shortcut, mf}. Then, the overall representation is used to condition each block as in DiT \citep{dit} or SiT \citep{sit}. Therefore, our length embedding is the only additional component which results in a negligible computational overhead.

We also introduce an overhead-free modification, where blocks can be combined into groups. Then, each group contains $G$ blocks and represents a single Euler step in $\textrm{ODE}_l$. For example, a sequence of $L$ blocks with $G=1$ is equivalent to (\ref{eq:ours}). When $G=2$, the number of available length-wise shortcuts is halved \ie~$L/2$. The conventional architecture without the inner $\textrm{ODE}_l$ is equivalent to the $G=L$ option. Therefore, we can empirically verify how our $\textrm{ODE}_l$ approach affects the sampling performance metrics by varying the group hyperparameter $G\in \{1,\ldots, L\}$. 

\section{Experiments}\label{sec:eval}
This section explores the quality-complexity tradeoff for various corner cases that have been enabled by our approach as illustrated in Fig. \ref{fig:arch}. In particular, we apply ODE$_t$(ODE$_l$) to the vanilla flow matching model \citep{liu2023flow} and complement recent NFE minimization methods \citep{shortcut, mf}. Also, we perform ablation studies on the fixed- vs. adaptive-step ODE solver perspective and introduced hyperparameters: finding the optimal group size $G$ and the hyperparameter $K$ for the proposed length consistency objective. Our code employs adaptive-step ODE solvers from the \textit{diffrax} library \citep{kidger2021on}.

\subsection{Experimental setup}\label{subsec:setup}
\textbf{Datasets and evaluation.} We experiment with CelebA-HQ \citep{celeba} for unconditional generation and ImageNet \citep{imagenet} for class-conditional generation. We use $256 \times 256$ image resolution in all experiments. After training, we sample images using the first-order Euler solver with 1, 4, and 128 discrete time steps $T$ in the outer ODE$_{t}$ following \citep{shortcut, mf} experiments. Additionally, we employ Dormand-Prince's 5/4 (Dopri5) adaptive-step solver \citep{dormand1980family} without first-order time discretization to highlight that the ODE$_{t}$ is solver-agnostic in our method. We also use classifier-free guidance \citep{ho2021classifierfree} and its \citep{mf} modification in the corresponding ImageNet sampling experiments and match the baseline hyperparameters. 

The quality of the generated images is evaluated using the Fréchet inception distance (FID) metric \citep{fid}. We report the FID-50K metric that is measured for 50,000 sampled images \wrt the corresponding dataset statistics. The metric is estimated using 2048-length latent activations by running the InceptionV3 neural network with public weights \citep{jax_fid}. Before processing activations, generated images are resized to $299 \times 299$ resolution with bilinear interpolation and clipped to the $ [-1, 1] $ range.

\textbf{Architectures and baselines.} We follow the \citep{shortcut, mf} setups and employ the DiT-B/XL \citep{dit} architectures which consist of a sequence with 12 and 28 transformer blocks, respectively. Additionally, we experiment with more recent SiT-B/XL~\citep{sit} networks with the same number of transformer blocks as in DiT. Importantly, the DiT and SiT architectures themselves operate in the latent space of the pretrained variational autoencoder (VAE) from \citet{rombach2021highresolution} with $8\times$ downscaling and upscaling in the encoder and decoder spatial resolutions, respectively.

A base DiT network without any compute optimizations has been trained using the variance preserving \citep{impr_ddpm} diffusion model (DM) \citep{dit} and the same network, but trained using the rectified flow matching (RFM) objective \citep{liu2023flow} from Section~\ref{sec:theory}. Then, we report reference results for distillation methods from Section \ref{sec:related} that have been reimplemented by \citet{shortcut} for the base RFM model. Distillation methods include two-stage methods with teacher-student training such as progressive distillation \citep{salimans2022} and Reflow from \citep{liu2023flow}, and single-stage or \textit{end-to-end methods} without explicit students \eg~the consistency training \citep{cm} and live Reflow \citep{shortcut}. The state-of-the-art NFE minimization methods \citep{shortcut, imm, mf} are reported in our Tables~\ref{tab:celeba-benchmark}-\ref{tab:imagenetxl-benchmark}, if the corresponding architecture and data set results are publicly available.

To show ODE$_t$(ODE$_l$) generality, we extend the \citep{shortcut} JAX code and \citep{mf} PyTorch code and adopt our approach while keeping their implementation and hyperparameters intact. Hence, our extension complements the baseline code with the steps described in Algs.~\ref{alg:1}-\ref{alg:2}. Then, we experiment with the ODE$_t$(ODE$_l$)-derived models as follows. We initialize all weights from scratch in the CelebA-HQ experiments. In case of large-scale ImageNet, we start from the publicly available checkpoints for \citep{shortcut, mf} to save training time and to show the feasibility of short finetuning. Next, we train all ODE$_t$(ODE$_l$) variants for $200,000$ iterations using the schedule and hyperparameters from the Appendix A. We report the FID-50K metric as $\mu_{\pm\sigma}$ where the mean $\mu$ and standard deviation error bars $\pm\sigma$ have been estimated using 4 runs with different random number generator seeds.

\begin{table}[t]
\caption{\textbf{CelebA-HQ-256 unconditional DiT-B generation.} The FID-50K image quality metric (lower is better), 
\%. The \fst{best} and the \scd{second best} results ($\mu_{\pm\sigma}$) are highlighted. Compute is proportional to the number of time steps in the Euler method, $T \in \{1,4,128\}$. In our ODE$_t$(ODE$_l$), compute and memory also decrease with the DiT-B neural network length, $l \in \{12, 8, 4\}$.
}
\label{tab:celeba-benchmark}
\centering
\resizebox{1.0\linewidth}{!}{
\begin{tabular}{lcccc}
\toprule
\multirow{2}{*}{\shortstack{Time steps,\\$T$}} & \multirow{2}{*}{\shortstack{End\\to end}} & \multicolumn{3}{c}{CelebA-HQ-256} \\
 & & 128 & 4 & 1 \\
\midrule
Base DM \citep{dit} & & 23.0 & (123) & (132) \\
Base RFM \citep{liu2023flow} & & 7.3 & 63.3 & (281) \\
Progr. dist. \citep{salimans2022}& \xmark & (303) & (251) & \fst{14.8} \\
Reflow \citep{liu2023flow} & \xmark & 16.1 & 18.4 & 23.2 \\
Cons. train. \citep{cm}& \cmark & 53.7 & 19.0 & 33.2 \\
Live Reflow \citep{shortcut} & \cmark & \fst{6.3} & 27.2 & 43.3 \\
SM \citep{shortcut}   & \cmark & 6.9 & \scd{13.8} & 20.5 \\
~+ ODE$_t$(ODE$_{l=12}$) & \cmark & \scd{6.6}\tiny$\pm$0.2 &\fst{12.7}\tiny$\pm$0.3 &\scd{17.3}\tiny$\pm$0.7 \\
~+ ODE$_t$(ODE$_{l= 8}$) & \cmark &      6.8\tiny$\pm$0.2  &\fst{12.7}\tiny$\pm$0.4 &17.7\tiny$\pm$0.7 \\
~+ ODE$_t$(ODE$_{l= 4}$) & \cmark &     14.3\tiny$\pm$0.7  &      15.7\tiny$\pm$0.5 &      31.8\tiny$\pm$0.9 \\
\bottomrule
\end{tabular}
}
\end{table}

\begin{table}[t]
\caption{\textbf{ImageNet-256 class-conditional DiT/SiT-B generation.} The FID-50K image quality metric (lower is better), 
\%. The \fst{best} and the \scd{second best} results ($\mu_{\pm\sigma}$) are highlighted. Compute is proportional to the number of time steps in the Euler method, $T \in \{1,4,128\}$. In our ODE$_t$(ODE$_l$), compute and memory also decrease with the DiT/SiT-B neural network length, $l \in \{12, 8\}$.
}
\label{tab:imagenetb-benchmark}
\centering
\resizebox{1.0\linewidth}{!}{
\begin{tabular}{lcccc}
\toprule
\multirow{2}{*}{\shortstack{Time steps,\\$T$}} & \multirow{2}{*}{\shortstack{End\\to end}} & \multicolumn{3}{c}{ImageNet-256} \\
& & 128 & 4 & 1 \\
\midrule
\multicolumn{5}{c}{DiT-B architecture:} \\
Base DM \citep{dit} & & 39.7 & (465) & (467) \\
Base RFM \citep{liu2023flow} & & 17.3 & (108) & (325) \\
Progr. dist. \citep{salimans2022}& \xmark & (202) & (143) & \scd{35.6} \\
Reflow \citep{liu2023flow} & \xmark & 16.9 & 32.8 & 44.8 \\
Cons. train. \citep{cm}& \cmark & 42.8 & 43.0 & 69.7 \\
Live Reflow \citep{shortcut} & \cmark & 46.3 & 95.8 & 58.1 \\
SM \citep{shortcut} & \cmark & \scd{15.5} & 28.3 & 40.3 \\
~+ ODE$_t$(ODE$_{l=12}$) & \cmark & \fst{11.9}\tiny$\pm$0.1 & \fst{13.3}\tiny$\pm$0.2 & \fst{18.3}\tiny$\pm$0.5 \\
~+ ODE$_t$(ODE$_{l= 8}$) & \cmark &       21.0\tiny$\pm$0.2 & \scd{20.5}\tiny$\pm$0.3 & 38.7\tiny$\pm$0.6 \\
\midrule 
\multicolumn{5}{c}{SiT-B architecture:} \\
Base DM \citep{sit} & & 33.5 & - & - \\ 
MF \citep{mf} & \cmark & \scd{5.67} & \fst{5.49} & \fst{6.17} \\
~+ ODE$_t$(ODE$_{l=12}$) & \cmark & \fst{5.65} & \scd{5.64} & \scd{6.65} \\
~+ ODE$_t$(ODE$_{l= 8}$) & \cmark & 9.41 & 9.82 & 15.55 \\
\bottomrule
\end{tabular}
}
\end{table}

\begin{table}[t]
\caption{\textbf{ImageNet-256 class-conditional DiT/SiT-XL generation.} The FID-50K image quality metric (lower is better), 
\%. The \fst{best} and the \scd{second best} results are highlighted. Compute is proportional to the number of time steps in the Euler method, $T \in \{1,4,128\}$. In our ODE$_t$(ODE$_l$), compute and memory also decrease with the DiT/SiT-XL neural network length, $l \in \{28, 20\}$.
}
\label{tab:imagenetxl-benchmark}
\centering
\resizebox{1.0\linewidth}{!}{
\begin{tabular}{lccccc}
\toprule
\multirow{2}{*}{\shortstack{Base models}} & \multirow{2}{*}{\shortstack{FID-50K}} & \multicolumn{2}{c}{\multirow{2}{*}{\shortstack{Time steps,\\$T$}}} & \multirow{2}{*}{\shortstack{Param.\\count}} & \multirow{2}{*}{\shortstack{Train.\\epochs}} \\
& & & & & \\
\midrule
ADM-G \citep{dhariwal2021diffusion} & 3.94 &  \multicolumn{2}{c}{250 $\times$ 2} & 554M & 400 \\
LDM-4-G \citep{rombach2021highresolution} & 3.60 &  \multicolumn{2}{c}{250 $\times$ 2} & 400M & 200 \\
DiT-XL \citep{dit} & 2.27 & \multicolumn{2}{c}{250 $\times$ 2} & 675M & 1,400 \\
SiT-XL \citep{sit} & 2.06 & \multicolumn{2}{c}{250 $\times$ 2} & 675M & 1,400 \\
\midrule
Efficient methods & \multicolumn{3}{c}{FID-50K} & \multirow{2}{*}{\shortstack{Param.\\count}} & \multirow{2}{*}{\shortstack{Train.\\epochs}} \\
Time steps, $T$ & 128 & 4 & 1 & & \\
\midrule
IMM DiT-XL \citep{imm}     &     - &  \fst{2.51} &  \scd{8.05} & 675M & - \\
MF DiT-XL \citep{mf}     & - & - & \fst{3.43} & 676M & 240 \\
SM DiT-XL \citep{shortcut} &  \fst{3.80} &  \fst{7.80} & 10.60 & 675M & 250 \\
~+ ODE$_t$(ODE$_{l=28}$)   &  \scd{3.98} &  8.01 &  9.49 & 676M & 43 \\
~+ ODE$_t$(ODE$_{l=20}$)   &  6.79 &  8.03 & 23.10 & 485M & 43 \\
\midrule
MF SiT-XL \citep{mf}     & 5.95 &       4.62       & \fst{3.40} & 676M & 240 \\
~+ ODE$_t$(ODE$_{l=28}$) & \scd{4.59} & \fst{3.94} & \scd{3.84} & 676M & 43 \\
~+ ODE$_t$(ODE$_{l=20}$) & \fst{4.23} & \scd{4.47} & 10.74 & 485M & 43 \\
\bottomrule
\end{tabular}
}
\end{table}

\subsection{Quantitative results}\label{subsec:quant}
Table~\ref{tab:celeba-benchmark} presents FID-50K results for CelebA-HQ-256 images generated using the low-complexity Euler solver with $T$ uniformly discretized time steps. The upper part contains the base DM \citep{dit} and RFM \citep{liu2023flow} results, which indicate the inability to sample acceptable images when the number of time steps is low. The two-stage methods \citep{salimans2022,liu2023flow} show a significant improvement when $T$ is low at the expense of a more complicated training scheme. The time shortcut modeling (SM) \citep{shortcut} has the best quality-complexity tradeoff because it achieves the lowest average FID-50K metrics both when $T$ is low and high. When our ODE$_t$(ODE$_l$) is applied to the SM, we can see that a shorter neural network with the length $l=8$ produces almost the same-quality images as its full-length version with $l=12$. Hence, the proposed method can further improve the quality-complexity tradeoff that is illustrated in Fig. \ref{fig:scaling} with $2\times$ latency scaling gain.

Tables~\ref{tab:imagenetb-benchmark}-\ref{tab:imagenetxl-benchmark} show the experimental results for ImageNet-256 class-conditional image generation using, correspondingly, smaller DiT/SiT-B networks with 12 transformer blocks and larger DiT/SiT-XL architectures with 28 blocks. Unlike the CelebA-HQ dataset with only 30,000 images, ImageNet contains more than 1.2 million images with class labels. Hence, it is a more challenging task to sample from a larger class-conditional data distribution while minimizing computational resources. Overall, we can see inferior FID-50K results in Table~\ref{tab:imagenetb-benchmark} compared to Table~\ref{tab:celeba-benchmark} with the small base DiT-B architecture. At the same time, our ODE$_t$(ODE$_l$) shows a significant improvement when added to the SM method, even with the smaller depth $l=8$ and NFE 1 or 4. On the other hand, the recent NFE minimization method MeanFlow \citep{mf} is capable of achieving substantially better FID-50K metrics even for a small SiT-B architecture using a combination of various techniques. In this case, our ODE$_t$(ODE$_l$) can match its state-of-the-art results with the full-length neural network and, additionally, minimize image quality degradation when operating using $2/3\times$ of the full depth ($l=8$) neural network.

Next, we experiment in Table~\ref{tab:imagenetxl-benchmark} with larger DiT/SiT-XL architectures, and check if the base model expressivity puts the upper bound on Table~\ref{tab:imagenetb-benchmark} metrics. First, we provide reference results for the base SiT-XL without sampling optimizations, which achieves 2.06 FID-50K using 500 sampling time steps compared to only 33.5 for SiT-B. Second, we observe a lack of consistency in results of the NFE minimization methods \citep{imm, shortcut, mf}. Surprisingly, none of the methods can claim the lowest FID in all time step cases $T \in \{1,4,128\}$. For example, the reference SM and our SM+ODE$_t$(ODE$_l$) variant perform the closest to the base model when $T=128$. At the same time, IMM \citep{imm} and MF \citep{mf} are the best when $T=4$ and $T=1$, respectively. Nonetheless, our ODE$_t$(ODE$_l$)-derived models either match or even outperform the corresponding NFE minimization baselines. Therefore, a NFE minimization method that performs consistently in all corner cases is yet to be discovered. Meanwhile, the proposed ODE$_t$(ODE$_l$) experiences no or minor degradation when operating with 30\% less parameters and latency \ie $l=20$ vs. $l=28$ blocks.

\begin{table}[t!]
\caption{\textbf{Ablation study: fixed- vs. adaptive-step solvers.} We compare FID, memory usage (MEM), and wall-clock time (WCT) to sample 128-size mini-batch using DiT-B for CelebA-HQ-256, sec. Compared to Euler, the adaptive-step Dopri5 is a better choice for high-quality sampling with low latency and memory usage.}
\label{tab:solver}
\centering
\resizebox{0.95\linewidth}{!}{
\begin{tabular}{lcccc}
    \toprule
    Method & ODE$_t$ solver & FID$\downarrow$ & MEM$\downarrow$ & WCT$\downarrow$ \\
    \midrule
    SM \citep{shortcut} &          Euler $(T=32)$ & 8.90 & $1\times$ & 2.09 \\
    ours, ${l=12}$ &          Euler $(T=32)$      & 8.23 & $1\times$ & 2.10 \\
    ours, ${l=12}$ & Dopri5 $(\textrm{tol.}~0.1)$ & \fst{6.57} & $1\times$ & \fst{1.84} \\
    \midrule
    ours, ${l= 8}$ & Euler $(T=32)$               & 8.25 & $2/3\times$ & 1.4 \\
    ours, ${l= 8}$ & Dopri5 $(\textrm{tol.}~0.1)$ & \fst{6.85} & $2/3\times$ & \fst{1.2} \\
    \midrule
    ours, ${l= 4}$ & Euler $(T=32)$               &  \fst{12.47} & $1/3\times$  & 0.7 \\
    ours, ${l= 4}$ & Dopri5 $(\textrm{tol.}~0.1)$ & 15.91 & $1/3\times$ & \fst{0.6} \\
    \bottomrule
\end{tabular}
}
\end{table}

\subsection{Ablation study}\label{subsec:ablation}
\textbf{Fixed- vs. adaptive-step solvers.} We study whether the common fixed-step Euler solver is optimal for all corner cases in Table~\ref{tab:solver}. We measure actual wall-clock time (WCT) latency to sample a mini-batch of images (excluding the VAE processing) for the Euler solver with $T$ time steps and adaptive-step Dopri5 solver with the reported tolerance (tol.) levels. We highlight that in our method both the WCT latency and the memory usage are linear with the neural network length and there is a negligible latency overhead for the additional length embedder \eg, 2.09 seconds for SM \citep{shortcut} vs. 2.10 seconds for ours with $l=12$. Table~\ref{tab:solver} shows that Dopri5 can sample images with high quality (lower FID) and significantly lower latency. For example, the SM method achieves 8.9 FID in 2.09 seconds, while our ODE$_t$(ODE$_{l=8}$) with Dopri5 and 0.1 tolerance reaches 6.85 FID in only 1.2 seconds. Hence, adaptive-step solvers are a better choice in the high-quality sampling regime, while the first-order Euler is only optimal for a few-step lower-quality generation as shown in Fig. \ref{fig:scaling} scaling curves.

\begin{table}
\caption{\textbf{Ablation study: group size $G$.} Larger group sizes lead to better FID results in the most compute-efficient regime (small $l$ and $T$), but provide less options for the test-time network length.}
\label{tab:group}
\centering
\begin{tabular}{lccc}
\toprule
Euler steps, $T$ & 128 & 4 & 1 \\
\midrule
$l=8, G=1$ & \scd{7.0} &      13.7  &      18.6  \\
$l=8, G=2$ &      7.3  & \fst{12.0} & \scd{17.8} \\
$l=8, G=4$ & \fst{6.8} & \scd{12.7} & \fst{17.7} \\
\midrule
$l=4, G=1$ &      17.9  &      17.0  &      35.8  \\
$l=4, G=2$ & \scd{16.6} & \scd{16.9} & \scd{32.9} \\
$l=4, G=4$ & \fst{14.3} & \fst{15.7} & \fst{31.8} \\
\bottomrule
\end{tabular}
\end{table}

\begin{table}
\caption{\textbf{Ablation study: length consistency $K$.} Higher consistency regularization using the hyperparameter $K$ helps to improve FID in the most compute-efficient configurations (small $l$ and $T$).}
\label{tab:consist}
\centering
\begin{tabular}{lccc}
\toprule
Euler steps, $T$ & 128 & 4 & 1 \\
\midrule
$l= 8, K=  0$ & \fst{6.7} &      12.9  & \scd{18.5} \\
$l= 8, K=1/8$ & \scd{6.8} & \scd{12.7} & \fst{17.7} \\
$l= 8, K=1/4$ &      6.9  & \fst{12.6} & \fst{17.7} \\
\midrule
$l= 4, K=  0$ & 15.9 & 15.9 & 31.9 \\
$l= 4, K=1/8$ & \scd{14.3} & \scd{15.7} & \scd{31.8} \\
$l= 4, K=1/4$ & \fst{13.2} & \fst{15.2} & \fst{30.6} \\
\bottomrule
\end{tabular}
\end{table}

\textbf{Group size $G$.} We perform ablation studies for key hyperparameters that accompany the framework proposed in Section~\ref{sec:prop} using the CelebA-HQ dataset. First, we vary the group size constant $G$ that defines how many steps are performed in the inner $\textrm{ODE}_l$ and, hence, the number of options for the neural network length. As expected, Table~\ref{tab:group} results show that the larger groups with higher expressivity lead to better (lower) FID scores. Surprisingly, it is not the case when both the depth $l$ and the number of time steps $T$ are high and $G=1$. This can be related to better gradient propagation during training with more residual connections.

\textbf{Length consistency $K$.} We investigate the hyperparameter $K$ that defines the share of length consistency terms in a mini-batch during training. According to Table~\ref{tab:consist} study, higher length consistency regularization is beneficial in the most compute-efficient regimes \ie~when $l$ and $T$ are low. We hypothesize that it sets how close a shorter-length implicit students follow the vector field of a full-length teacher.

\subsection{Qualitative results and complexity scaling}\label{subsec:qual}
\begin{figure}
\centering
\includegraphics[width=0.8\columnwidth]{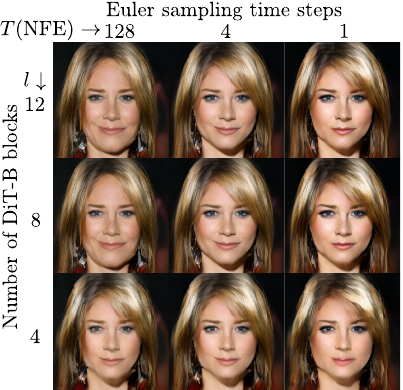}
\caption{\textbf{DiT-B CelebA sampling.} Time shortcuts (SM) \citep{shortcut} (first row) alter the image style, our length shortcuts (columns) preserve the style and iteratively compress the image details.}
\label{fig:qual}
\end{figure}

\begin{figure}
\centering
\includegraphics[width=0.82\columnwidth]{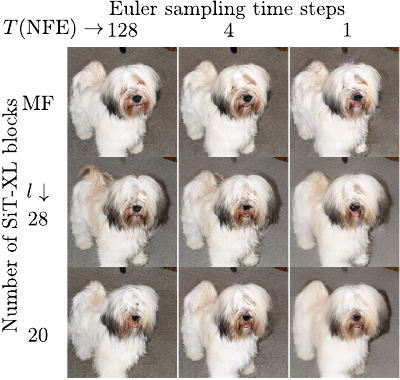}
\caption{\textbf{SiT-XL ImageNet sampling.} Compared to MF \citep{mf} (first row), our ODE$_t$(ODE$_{l=20}$) with 30\% less compute in the last row contains moderately less semantic details when NFE$=1$.}
\label{fig:sitxl-qual}
\end{figure}

\begin{figure}
\centering
\includegraphics[width=0.85\columnwidth]{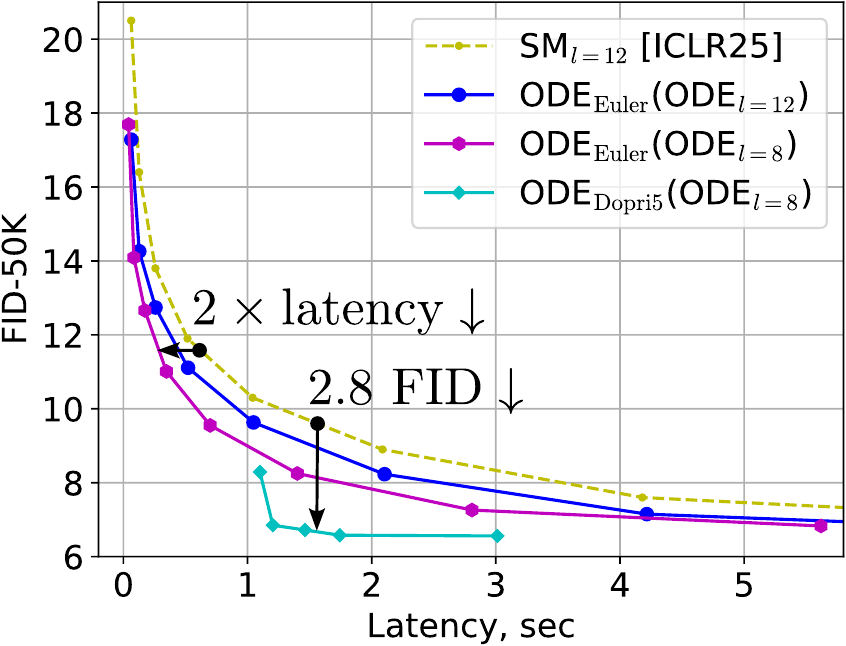}
\caption{\textbf{CelebA FID vs. DiT-B latency.} ODE$_t$(ODE$_l$) scales better than \citep{shortcut}: it shows up to $2\times$ reduction in latency with Euler solver in compute-optimized mode and provides up to $2.8$ lower FID score with adaptive-step solver for high-quality sampling.}
\label{fig:scaling}
\end{figure}

Our representative sampling results for CelebA and ImageNet are shown in Figures~\ref{fig:qual}-\ref{fig:sitxl-qual} and using the $3\times3$ image grid. We decrease generation time steps $T$ horizontally from 128 to only a single step, and the network depth $l$ vertically from the full 12 blocks to 4 blocks and from 28 blocks to 20 blocks for DiT-B and SiT-XL, respectively. The upper row in Fig.~\ref{fig:qual} contains the SM \citep{shortcut} time shorcuts only, where we can see noticeable changes in the image style caused by the time consistency objective. On the other hand, our length shortcuts with the corresponding consistency objective from (\ref{eq:ours}) preserve the style and iteratively compress the image details. We explain this phenomenon as follows: our objective explicitly constrains neural network to follow the same ODE path, while the \citep{shortcut} time consistency only implicitly regularizes the ODE trajectory as two half-size time steps which leads to Fig.~\ref{fig:qual} artifacts. The bottom right corner combines both approaches with the benefit of extremely fast sampling ($400\times$ lower latency \wrt the top left corner). We explicitly illustrate the quality-complexity scaling in Fig.~\ref{fig:scaling}.

ImageNet examples are shown in Fig.~\ref{fig:sitxl-qual} where the upper row with the MF \citep{mf} baseline is similar in quality to our full-length ODE$_t$(ODE$_{l=28}$) that is justified by quantitative metrics in Table~\ref{tab:imagenetxl-benchmark}. On the other hand, ODE$_t$(ODE$_{l=20}$) with 30\% less compute has almost the same visual quality when $T>1$ and iteratively less semantic details with a few-step generation \eg, $T=1$ in bottom right corner. We provide more qualitative examples in the Appendix~B.

\section{Conclusion}\label{sec:conc}
In this paper, we presented the ODE$_t$(ODE$_l$) that enabled shortcuts in the neural network depth dimension. Our approach is analytically grounded on the observation that recent transformer-based architectures consist of a sequence of blocks that can solve the inner ODE$_l$ with variable depth.

To implement the inner ODE$_l$, we proposed novel length consistency training objective and architectural modifications with negligible computational overhead. In general, our ODE$_t$(ODE$_l$) can be used separately or can complement NFE minimization methods for faster ODE-based sampling using flow and diffusion models. CelebA-HQ and ImageNet generation experiments have shown further improvement in the quality-complexity tradeoff scaling. Importantly, it allowed to mitigate memory bottleneck in large-scale models and, as a future work, to adaptively select computational resources by modulating the length variable. 

{\small
\bibliographystyle{ieeenat_fullname}
\bibliography{paper}}

@String(CVPR= {IEEE Conf. Comput. Vis. Pattern Recog.})

@String(ICCV= {Int. Conf. Comput. Vis.})

@String(ECCV= {Eur. Conf. Comput. Vis.})

@String(NIPS= {Adv. Neural Inform. Process. Syst.})

@String(ICLR = {Int. Conf. Learn. Represent.})

@String(ICML = {ICML})

@String(IJCAI = {IJCAI})

@String(CVPR  = {CVPR})

@String(ICCV  = {ICCV})

@String(ECCV  = {ECCV})

@String(NIPS  = {NeurIPS})

@String(ICLR  = {ICLR})

@String(ICML  = {ICML})

@inproceedings{lipman2023flow,
	title={Flow Matching for Generative Modeling},
	author={Yaron Lipman and Ricky T. Q. Chen and Heli Ben-Hamu and Maximilian Nickel and Matthew Le},
	booktitle=ICLR,
	year={2023}
}

@inproceedings{rombach2021highresolution,
	author    = {Robin Rombach and Andreas Blattmann and Dominik Lorenz and Patrick Esser and Björn Ommer},
	title     = {High-Resolution Image Synthesis with Latent Diffusion Models},
	booktitle = CVPR,
	year      = {2022}
}

@inproceedings{esser2024scaling,
	title={Scaling Rectified Flow Transformers for High-Resolution Image Synthesis},
	author={Patrick Esser and Sumith Kulal and Andreas Blattmann and Rahim Entezari and Jonas M{\"u}ller and Harry Saini and Yam Levi and Dominik Lorenz and Axel Sauer and Frederic Boesel and Dustin Podell and Tim Dockhorn and Zion English and Robin Rombach},
	booktitle=ICML,
	year={2024},
}

@inproceedings{sohl2015diffusion,
	author    = {Jascha Sohl{-}Dickstein and Eric A. Weiss and Niru Maheswaranathan and Surya Ganguli},
	title     = {Deep Unsupervised Learning using Nonequilibrium Thermodynamics},
	booktitle = ICML,
	year      = {2015}
}

@article{tong2024improving,
	title={Improving and generalizing flow-based generative models with minibatch optimal transport},
	author={Alexander Tong and Kilian Fatras and Nikolay Malkin and Guillaume Huguet and Yanlei Zhang and Jarrid Rector-Brooks and Guy Wolf and Yoshua Bengio},
	journal={Transactions on Machine Learning Research},
	year={2024},
}

@inproceedings{albergo2023building,
	title={Building Normalizing Flows with Stochastic Interpolants},
	author={Michael Samuel Albergo and Eric Vanden-Eijnden},
	booktitle=ICLR,
	year={2023},
}

@inproceedings{aca,
	title = {Adaptive Checkpoint Adjoint Method for Gradient Estimation in Neural {ODE}},
	author = {Zhuang, Juntang and Dvornek, Nicha and Li, Xiaoxiao and Tatikonda, Sekhar and Papademetris, Xenophon and Duncan, James},
	booktitle = ICML,
	year = {2020},
}

@inproceedings{anode,
	title={{ANODE}: Unconditionally Accurate Memory-Efficient Gradients for Neural {ODE}s},
	author={Amir Gholami and Kurt Keutzer and George Biros},
	booktitle=IJCAI,
	year={2019},
}

@inproceedings{song_generative_2019,
	title        = {Generative Modeling by Estimating Gradients of the Data Distribution},
	author       = {Song, Yang and Ermon, Stefano},
	year         = 2019,
	booktitle    = NIPS
}

@inproceedings{ho_denoising_2020,
	title        = {Denoising Diffusion Probabilistic Models},
	author       = {Ho, Jonathan and Jain, Ajay and Abbeel, Pieter},
	year         = 2020,
	booktitle    = NIPS
}

@article{liu_rectified_2022,
	title        = {Rectified Flow: A Marginal Preserving Approach to Optimal Transport},
	author       = {Liu, Qiang},
	year         = 2022,
	journal      = {arXiv:2209.14577}
}

@inproceedings{liu2023flow,
	title={Flow Straight and Fast: Learning to Generate and Transfer Data with Rectified Flow},
	author={Xingchao Liu and Chengyue Gong and Qiang Liu},
	booktitle=ICLR,
	year={2023},
}

@inproceedings{ffjord,
	title={Scalable Reversible Generative Models with Free-form Continuous Dynamics},
	author={Will Grathwohl and Ricky T. Q. Chen and Jesse Bettencourt and David Duvenaud},
	booktitle=ICLR,
	year={2019},
}

@inproceedings{dit,
	author    = {Peebles, William and Xie, Saining},
	title     = {Scalable Diffusion Models with Transformers},
	booktitle = ICCV,
	year      = {2023},
}

@inproceedings{geffner2025proteina,
	title={Proteina: Scaling Flow-based Protein Structure Generative Models},
	author={Tomas Geffner and Kieran Didi and Zuobai Zhang and Danny Reidenbach and Zhonglin Cao and Jason Yim and Mario Geiger and Christian Dallago and Emine Kucukbenli and Arash Vahdat and Karsten Kreis},
	booktitle=ICLR,
	year={2025},
}

@inproceedings{shortcut,
	title={One Step Diffusion via Shortcut Models},
	author={Kevin Frans and Danijar Hafner and Sergey Levine and Pieter Abbeel},
	booktitle=ICLR,
	year={2025},
}

@inproceedings{bar2024lumiere,
	title={Lumiere: A space-time diffusion model for video generation},
	author={Omer Bar-Tal and Hila Chefer and Omer Tov and Charles Herrmann and Roni Paiss and Shiran Zada and Ariel Ephrat and Junhwa Hur and Guanghui Liu and Amit Raj and Yuanzhen Li and Michael Rubinstein and Tomer Michaeli and Oliver Wang and Deqing Sun and Tali Dekel and Inbar Mosseri},
	booktitle={SIGGRAPH Asia},
	year={2024}
}

@inproceedings{sahoo2024simple,
	title={Simple and Effective Masked Diffusion Language Models},
	author={Subham Sekhar Sahoo and Marianne Arriola and Aaron Gokaslan and Edgar Mariano Marroquin and Alexander M Rush and Yair Schiff and Justin T Chiu and Volodymyr Kuleshov},
	booktitle=NIPS,
	year={2024},
}

@inproceedings{omniflow,
	title={Omni{F}low: Any-to-Any Generation with Multi-Modal Rectified Flows}, 
	author={Shufan Li and Konstantinos Kallidromitis and Akash Gokul and Zichun Liao and Yusuke Kato and Kazuki Kozuka and Aditya Grover},
	booktitle = CVPR,
	year={2025},
}

@misc{dieleman2024distillation,
	author = {Dieleman, Sander},
	title = {The paradox of diffusion distillation},
	url = {https://sander.ai/2024/02/28/paradox.html},
	year = {2024}
}

@inproceedings{zhu2025slimflow,
	title={Slim{F}low: Training Smaller One-Step Diffusion Models with Rectified Flow},
	author={Zhu, Yuanzhi and Liu, Xingchao and Liu, Qiang},
	booktitle=ECCV,
	year={2025},
}

@article{hinton2015,
	title={Distilling the Knowledge in a Neural Network}, 
	author={Geoffrey Hinton and Oriol Vinyals and Jeff Dean},
	year={2015},
	journal={arXiv:1503.02531},
}

@inproceedings{schuster2021,
	title = "Consistent Accelerated Inference via Confident Adaptive Transformers",
	author = "Schuster, Tal  and
	Fisch, Adam  and
	Jaakkola, Tommi  and
	Barzilay, Regina",
	editor = "Moens, Marie-Francine  and
	Huang, Xuanjing  and
	Specia, Lucia  and
	Yih, Scott Wen-tau",
	booktitle = "Proceedings of the Conference on Empirical Methods in Natural Language Processing",
	year = "2021",
}

@inproceedings{MoonCYYLC024,
	author={Tae Hong Moon and Moonseok Choi and EungGu Yun and Jongmin Yoon and Gayoung Lee and Jaewoong Cho and Juho Lee},
	title={A Simple Early Exiting Framework for Accelerated Sampling in Diffusion Models},
	year={2024},
	booktitle=ICML,
}

@inproceedings{zhao2025dynamic,
	title={Dynamic Diffusion Transformer},
	author={Wangbo Zhao and Yizeng Han and Jiasheng Tang and Kai Wang and Yibing Song and Gao Huang and Fan Wang and Yang You},
	booktitle=ICLR,
	year={2025},
}

@inproceedings{you2025layer,
	title={Layer-and Timestep-Adaptive Differentiable Token Compression Ratios for Efficient Diffusion Transformers},
	author={You, Haoran and Barnes, Connelly and Zhou, Yuqian and Kang, Yan and Du, Zhenbang and Zhou, Wei and Zhang, Lingzhi and Nitzan, Yotam and Liu, Xiaoyang and Lin, Zhe and Shechtman, Eli and Amirghodsi, Sohrab and Lin, Yingyan Celine},
	booktitle=CVPR,
	year={2025}  
}

@article{luhman2021knowledge,
	title={Knowledge Distillation in Iterative Generative Models for Improved Sampling Speed}, 
	author={Eric Luhman and Troy Luhman},
	year={2021},
	journal={arXiv:2101.02388},
}

@inproceedings{salimans2022,
	title={Progressive Distillation for Fast Sampling of Diffusion Models},
	author={Tim Salimans and Jonathan Ho},
	booktitle=ICLR,
	year={2022},
}

@inproceedings{Meng_2023_CVPR,
	author    = {Meng, Chenlin and Rombach, Robin and Gao, Ruiqi and Kingma, Diederik and Ermon, Stefano and Ho, Jonathan and Salimans, Tim},
	title     = {On Distillation of Guided Diffusion Models},
	booktitle = CVPR,
	year      = {2023},
}

@inproceedings{song2024improved,
	title={Improved Techniques for Training Consistency Models},
	author={Yang Song and Prafulla Dhariwal},
	booktitle=ICLR,
	year={2024},
}

@inproceedings{geng2025consistency,
	title={Consistency Models Made Easy},
	author={Zhengyang Geng and Ashwini Pokle and Weijian Luo and Justin Lin and J Zico Kolter},
	booktitle=ICLR,
	year={2025},
}

@inproceedings{cm,
	title = {Consistency Models},
	author = {Song, Yang and Dhariwal, Prafulla and Chen, Mark and Sutskever, Ilya},
	booktitle = ICML,
	year = 	 {2023},
}

@inproceedings{Bai_2024_CVPR,
	author    = {Bai, Xingjian and Melas-Kyriazi, Luke},
	title     = {Fixed Point Diffusion Models},
	booktitle = CVPR,
	year      = {2024},
}

@inproceedings{Zhang_2024_CVPR,
	author    = {Zhang, Junyu and Liu, Daochang and Park, Eunbyung and Zhang, Shichao and Xu, Chang},
	title     = {Residual Learning in Diffusion Models},
	booktitle = CVPR,
	year      = {2024},
}

@inproceedings{ma2024neural,
	title={Neural Residual Diffusion Models for Deep Scalable Vision Generation},
	author={Zhiyuan Ma and Liangliang Zhao and Biqing Qi and Bowen Zhou},
	booktitle=NIPS,
	year={2024},
}

@inproceedings{Wu_2018_CVPR,
	author = {Wu, Zuxuan and Nagarajan, Tushar and Kumar, Abhishek and Rennie, Steven and Davis, Larry S. and Grauman, Kristen and Feris, Rogerio},
	title = {Block{D}rop: Dynamic Inference Paths in Residual Networks},
	booktitle = CVPR,
	year = {2018}
}

@inproceedings{bengio2016,
	title={Conditional Computation in Neural Networks for faster models}, 
	author={Emmanuel Bengio and Pierre-Luc Bacon and Joelle Pineau and Doina Precup},
	booktitle=ICLR,
	year={2016},
}

@inproceedings{Wang_2018_ECCV,
	author = {Wang, Xin and Yu, Fisher and Dou, Zi-Yi and Darrell, Trevor and Gonzalez, Joseph E.},
	title = {Skip{N}et: Learning Dynamic Routing in Convolutional Networks},
	booktitle = ECCV,
	year = {2018}
}

@inproceedings{beyond,
	title = 	 {Beyond Finite Layer Neural Networks: Bridging Deep Architectures and Numerical Differential Equations},
	author =       {Lu, Yiping and Zhong, Aoxiao and Li, Quanzheng and Dong, Bin},
	booktitle = 	 ICML,
	year = 	 {2018},
}

@inproceedings{song2021scorebased,
	title={Score-Based Generative Modeling through Stochastic Differential Equations},
	author={Yang Song and Jascha Sohl-Dickstein and Diederik P Kingma and Abhishek Kumar and Stefano Ermon and Ben Poole},
	booktitle=ICLR,
	year={2021},
}

@article{luo2023latent,
	title={Latent Consistency Models: Synthesizing High-Resolution Images with Few-Step Inference}, 
	author={Simian Luo and Yiqin Tan and Longbo Huang and Jian Li and Hang Zhao},
	journal={arXiv:2310.04378},
	year={2023},
}

@inproceedings{bai2019deep,
	title={Deep equilibrium models},
	author={Bai, Shaojie and Kolter, J Zico and Koltun, Vladlen},
	booktitle=NIPS,
	year={2019}
}

@misc{torchdeq,
	author = {Zhengyang Geng and J. Zico Kolter},
	title = {TorchDEQ: A Library for Deep Equilibrium Models},
	year = {2023},
	publisher = {GitHub},
	journal = {GitHub repository},
	howpublished = {\url{https://github.com/locuslab/torchdeq}},
}

@inproceedings{pdenet,
	title = 	 {{PDE}-Net: Learning {PDE}s from Data},
	author =       {Long, Zichao and Lu, Yiping and Ma, Xianzhong and Dong, Bin},
	booktitle = 	 ICML,
	year = 	 {2018},
}

@inproceedings{bilos,
	title={Neural Flows: Efficient Alternative to Neural {ODE}s},
	author={Marin Bilo{\v{s}} and Johanna Sommer and Syama Sundar Rangapuram and Tim Januschowski and Stephan G{\"u}nnemann},
	booktitle=NIPS,
	year={2021},
}

@inproceedings{zhang2023fast,
	title={Fast Sampling of Diffusion Models with Exponential Integrator},
	author={Qinsheng Zhang and Yongxin Chen},
	booktitle=ICLR,
	year={2023},
}

@inproceedings{ho2021classifierfree,
title={Classifier-Free Diffusion Guidance},
author={Jonathan Ho and Tim Salimans},
booktitle={NeurIPS 2021 Workshop on Deep Generative Models and Downstream Applications},
year={2021},
}

@inproceedings{gu2023boot,
title={{BOOT}: Data-free Distillation of Denoising Diffusion Models with Bootstrapping},
author={Jiatao Gu and Shuangfei Zhai and Yizhe Zhang and Lingjie Liu and Joshua M. Susskind},
booktitle={ICML 2023 Workshop on Structured Probabilistic Inference {\&} Generative Modeling},
year={2023},
}

@phdthesis{kidger2021on,
    title={{O}n {N}eural {D}ifferential {E}quations},
    author={Patrick Kidger},
    year={2021},
    school={University of Oxford},
}

@inproceedings{imagenet,
AUTHOR = {Deng, J. and Dong, W. and Socher, R. and Li, L.-J. and Li, K. and Fei-Fei, L.},
TITLE = {{Image{N}et: A Large-Scale Hierarchical Image Database}},
BOOKTITLE = CVPR,
YEAR = {2009},
}

@inproceedings{celeba,
title={Progressive Growing of {GAN}s for Improved Quality, Stability, and Variation},
author={Tero Karras and Timo Aila and Samuli Laine and Jaakko Lehtinen},
booktitle=ICLR,
year={2018},
}

@article{dormand1980family,
    author={Dormand, J. R. and Prince, P. J.},
    title={A family of embedded {R}unge--{K}utta formulae},
    journal={J. Comp. Appl. Math},
    year={1980},
}

@inproceedings{fid,
  title={{GAN}s trained by a two time-scale update rule converge to a local nash equilibrium},
  author={Heusel, Martin and Ramsauer, Hubert and Unterthiner, Thomas and Nessler, Bernhard and Hochreiter, Sepp},
  booktitle=NIPS,
  year={2017}
}

@misc{jax_fid,
author = {Wright, Matthias},
title = {{FID} computation in {JAX}/{F}lax},
url = {https://github.com/matthias-wright/jax-fid/},
year={2021}
}

@inproceedings{impr_ddpm,
  title = {Improved Denoising Diffusion Probabilistic Models},
  author = {Nichol, Alexander Quinn and Dhariwal, Prafulla},
  booktitle = ICML,
  year = 	 {2021},
}

@inproceedings{dhariwal2021diffusion,
title={Diffusion Models Beat {GAN}s on Image Synthesis},
author={Prafulla Dhariwal and Alexander Quinn Nichol},
booktitle=NIPS,
year={2021},
}

@inproceedings{zheng2023dpmsolverv,
title={{DPM}-Solver-v3: Improved Diffusion {ODE} Solver with Empirical Model Statistics},
author={Kaiwen Zheng and Cheng Lu and Jianfei Chen and Jun Zhu},
booktitle=NIPS,
year={2023},
}

@inproceedings{Zhou_2024_CVPR,
author    = {Zhou, Zhenyu and Chen, Defang and Wang, Can and Chen, Chun},
title     = {Fast {ODE}-based Sampling for Diffusion Models in Around 5 Steps},
booktitle = CVPR,
year      = {2024},
}

@inproceedings{chen2024on,
title={On the Trajectory Regularity of {ODE}-based Diffusion Sampling},
author={Defang Chen and Zhenyu Zhou and Can Wang and Chunhua Shen and Siwei Lyu},
booktitle=ICML,
year={2024},
}

@inproceedings{frankel2025ss,
title={{S4S}: Solving for a Fast Diffusion Model Solver},
author={Eric Frankel and Sitan Chen and Jerry Li and Pang Wei Koh and Lillian J. Ratliff and Sewoong Oh},
booktitle=ICML,
year={2025},
}

@inproceedings{imm,
title={Inductive Moment Matching},
author={Linqi Zhou and Stefano Ermon and Jiaming Song},
booktitle=ICML,
year={2025}
}

@inproceedings{mf,
title={Mean Flows for One-step Generative Modeling}, 
author={Zhengyang Geng and Mingyang Deng and Xingjian Bai and J. Zico Kolter and Kaiming He},
booktitle=NIPS,
year={2025}, 
}

@inproceedings{sit,
author="Ma, Nanye and Goldstein, Mark and Albergo, Michael S. and Boffi, Nicholas M. and Vanden-Eijnden, Eric and Xie, Saining",
title="{SiT}: Exploring Flow and Diffusion-Based Generative Models with Scalable Interpolant Transformers",
booktitle = ECCV,
year = {2024},
}

@article{shen2025efficient,
title={Efficient Diffusion Models: A Survey},
author={Hui Shen and Jingxuan Zhang and Boning Xiong and Rui Hu and Shoufa Chen and Zhongwei Wan and Xin Wang and Yu Zhang and Zixuan Gong and Guangyin Bao and Chaofan Tao and Yongfeng Huang and Ye Yuan and Mi Zhang},
journal={Transactions on Machine Learning Research},
year={2025},
}

@inproceedings{wang2025differentiable,
title={Differentiable Solver Search for Fast Diffusion Sampling},
author={Shuai Wang and Zexian Li and Qipeng zhang and Tianhui Song and Xubin Li and Tiezheng Ge and Bo Zheng and Limin Wang},
booktitle=ICML,
year={2025},
}
\onecolumn
\appendix

\section{Implementation details}\label{sec:app_impl}
Table~\ref{tab:hyper} shows the implementation details for DiT experiments in JAX. Typical DiT-B training run with 200,000 iterations takes 2 days using A6000 GPUs, while DiT-XL training takes 1 day using H100 GPUs. Hyperparameters and checkpoints for experiments with SiT follow the unofficial MeanFlow reimplementation in PyTorch\footnote{\href{https://github.com/zhuyu-cs/MeanFlow}{MeanFlow reimplementation at github.com/zhuyu-cs/MeanFlow}}. All SiT models have been trained with 200,000 iterations from the provided checkpoints using 1e-4 learning rate.

\begin{table*}[ht]
\caption{\textbf{Implementation details and hyperparameters.}}
\label{tab:hyper}
\centering
\begin{tabular}{lccc}
    \toprule
    Dataset & CelebA-HQ-256 & \multicolumn{2}{c}{ImageNet-256} \\
    \midrule
    Architecture & DiT-B & DiT-B & DiT-XL \\
    Patch size  & $2\times2$ & $2\times2$ & $2\times2$ \\
    Hidden size  & 768 & 768 & 1152 \\
    Attention heads & 12 & 12 & 16 \\
    MLP hidden ratio & 4 & 4 & 4 \\
    Blocks $\{L_\textrm{min},  L\}$ & $\{4,12\}$ & $\{4,12\}$ & $\{12,28\}$ \\
    Default $G,K$ & $4, 1/8$ & $4, 1/8$ & $8, 1/8$\\
    \midrule
    Training iterations & \multicolumn{3}{c}{$100K\rightarrow100K$}\\
    Learning rate & \multicolumn{3}{c}{1e-4~$\rightarrow$~1e-5} \\
    Batch size & 128 & 256 & 256 \\
    Initial weights & from scratch & \multicolumn{2}{c}{from \citep{shortcut} checkpoint}\\
    Schedule & \multicolumn{3}{c}{90\% const. with 5,000 iter. warmup $\rightarrow$ 10\% cosine decay} \\
    Optimizer & \multicolumn{3}{c}{AdamW with $\beta_1=0.9$, $\beta_2=0.999$ and weight decay $=0.1$} \\
    EMA ratio & \multicolumn{3}{c}{0.999}\\
    Class. free guid. & 0 & 1.5 & 1.5 \\
    GPUs & $4\times$A6000 & $8\times$A6000 & $8\times$H100 \\
    \midrule
    Generation & uncond. & class-cond. & class-cond. \\
    Classes & 1 & 1000 & 1000 \\
    $\textrm{ODE}_l$ solver & \multicolumn{3}{c}{Euler by the neural network} \\
    $\textrm{ODE}_t$ solver & Euler/Dopri5 & Euler & Euler \\
    EMA weights & \cmark & \cmark & \cmark \\
    \bottomrule
\end{tabular}
\end{table*}

\begin{figure*}[t]
\centering
\includegraphics[width=0.8\linewidth]{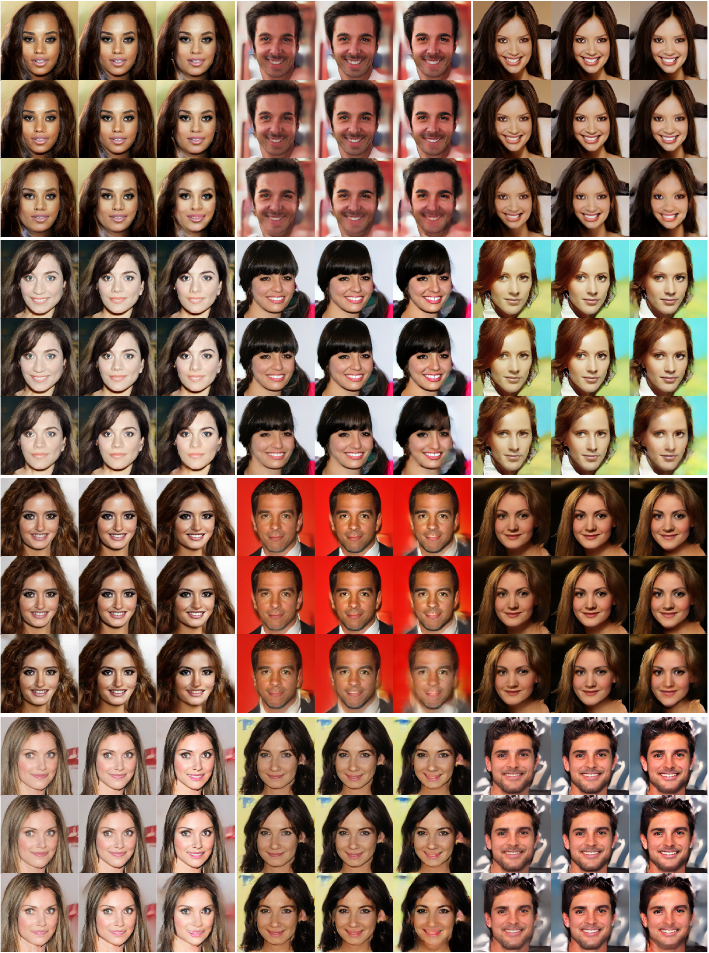}
\caption{CelebA-HQ-256 images generated by ODE$_t$(ODE$_l$) with DiT-B. Time shortcuts \citep{shortcut} ($T=128,4,1$ from left to right) alter the image style, our length shortcuts ($l=12,8,4$ from top to bottom) preserve the style and iteratively compress the image details. The bottom right corner combines both approaches with artifacts of two types and the benefit of extremely fast sampling with a factor of $400\times$ lower latency \wrt the top left corner.}\label{fig:app_qual1}
\end{figure*}

\begin{figure*}[t]
\centering
\includegraphics[width=0.8\linewidth]{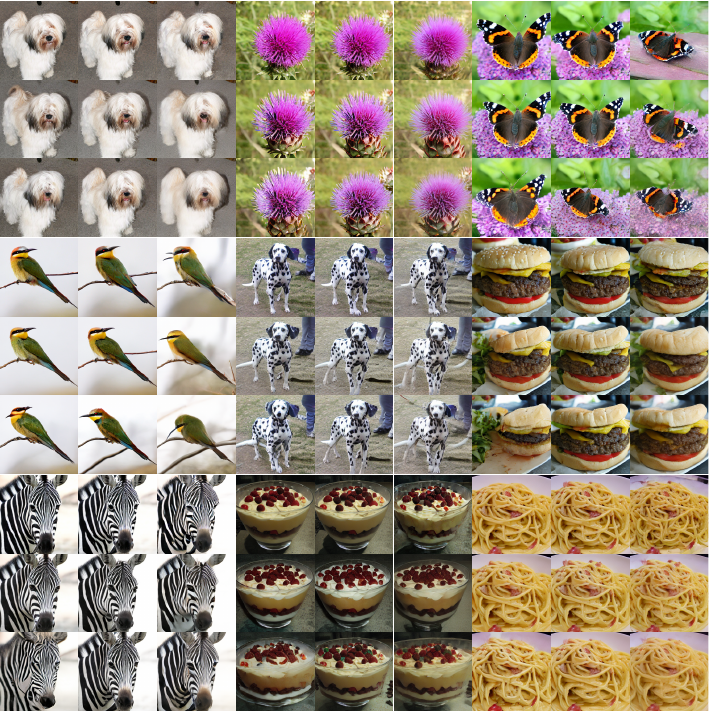}
\caption{ImageNet-256 images generated by the MF \citep{mf} and ODE$_t$(ODE$_l$) with SiT-XL. Euler steps $T=128,4,1$ decrease from left to right. Top row is the MF \citep{mf} baseline, while our ODE$_t$(ODE$_l$) is showed in the second row with $l=28$ and the last row with $l=20$. Our ODE$_t$(ODE$_{l=20}$) provides competitive quality when $T>1$ (NFE) at 30\% lower computational complexity.}
\label{fig:app_qual2}
\end{figure*}

\section{Qualitative results}\label{sec:app_qual}
Our additional qualitative sampling results for CelebA-HQ and ImageNet-1K with DiT-B and SiT-XL architectures are shown in Figures~\ref{fig:app_qual1}-\ref{fig:app_qual2} using $3\times3$ image grids. The generation time steps $T=128,4,1$ decrease horizontally from left to right. The neural network depth decreases vertically from top to bottom with $l=12,8,4$ for DiT-B and~$l=28,20$ for SiT-XL. 


\end{document}